\documentclass[runningheads]{llncs}
\usepackage[T1]{fontenc}
\usepackage{graphicx}
\usepackage{amsmath}
\usepackage{amssymb}
\usepackage{float}
\usepackage{microtype}
\usepackage{booktabs} 
\usepackage{marvosym}

\begin{document}
\title{Beyond Parallel Tracking: Interactive Multi-Feature Fusion Drives Semantic Reconstruction from Non-invasive Brain Recordings}
\titlerunning{Interactive Multi-Feature Fusion Drives Semantic Reconstruction}
\author{Boda Xiao\inst{1,2} \and
Xiran Xu\inst{2,3}\and 
Songyi Li\inst{2,3}\and
Yujie Yan\inst{1,2}\and
Xihong Wu\inst{2,3}\and
Heping Cheng\inst{4}\and
Jing Chen\inst{2,3,4}\textsuperscript{(\Letter)}}

\authorrunning{B. Xiao et al.}
\institute{Center for BioMed-X Research, Academy for Advanced Interdisciplinary Studies,Peking University, Beijing, China \and
Speech and Hearing Research Center, School of Intelligence Science and Technology,Peking University, Beijing, China\and
State Key Laboratory of General Artificial Intelligence, Beijing, China\and
National Biomedical Imaging Center, State Key Laboratory of Membrane Biology,
Institute of Molecular Medicine, Peking-Tsinghua Center for Life Sciences, 
College of Future Technology, Peking University, Beijing, China\\
\email{janechenjing@pku.edu.cn}}

\maketitle 
\begin{abstract}
Continuous semantic reconstruction from non-invasive neural recordings remains limited by the representational mismatch between semantic feature spaces and neural coding patterns, which severely impedes cross-modal alignment between high-noise neural signals and target semantic features. Prior semantic decoders have predominantly relied on static lexical representations or dynamic contextualized representations in isolation. This single-feature approach inevitably leads to severe information loss, as it fails to account for the human brain's capacity to integrate stable word attributes and dynamic contexts simultaneously. 

To bridge this gap, this study introduces a multi-feature fusion framework for non-invasive semantic reconstruction, systematically benchmarking two integration approaches: linear Naive Concatenation and non-linear Multi-Head Cross-Attention. Within this framework, our approach complements static lexical representations (\text{W2V}) with dynamic contextual representations (\text{GPT}) via an interactive gating mechanism to facilitate cooperative processing during language comprehension. 

Evaluated through extensive semantic reconstruction and text generation experiments, our framework reveals a robust performance hierarchy: $\text{Cross-Att} > \text{Concat} > \text{GPT} > \text{W2V}$. Crucially, the non-linear cross-attention fusion method achieves state-of-the-art performance, demonstrating that neural language decoding benefits from simulating the collaborative modulation between contextual information and core lexical attributes rather than depending on isolated individual features, while also offering a viable non-invasive brain-to-text decoding method.

\end{abstract}

\section{Introduction}
\subsection{The Cross-Modal Alignment Challenge in Semantic Reconstruction}
Semantic reconstruction aims to translate neural signals into continuous natural text \cite{anumanchipalli2019speech,tang2023semantic}. While speech decoders trained on invasive neural signals \cite{card2024accurate,metzger2023high,willett2023high} leverage high-SNR neural signals \cite{duraivel2023high,li2025high} to yield robust results, their widespread usability is constrained by significant surgical risks. Consequently, decoders based on safer, non-invasive neural signals like magnetoencephalography (MEG) and electroencephalography (EEG) have emerged as scalable alternatives. Prior efforts have explored the mapping of these non-invasive neural recordings onto discrete semantic elements \cite{dascoli2025towards} or the benchmarking of representational differences \cite{xiao2024comparing}. Concurrently, continuous language reconstruction has been demonstrated via fMRI voxels \cite{tang2023semantic} and investigated using MEG-based semantic decoders \cite{bowang2024icassp,bowang2024ieee}, while the EEG modality has been introduced for auditory attention tracking, spectrogram reconstruction, and acoustic speech reconstruction \cite{xu2024icassp,xu2024icasspw,defossez2023decoding}, but using EEG for continuous text reconstruction remains extremely challenging.

Despite these advances, non-invasive semantic reconstruction suffers from a profound bottleneck: the representational mismatch between semantic feature spaces and neural coding patterns, which impedes accurate cross-modal alignment. Traditional frameworks mitigate this by regressing a single semantic feature, relying either on static lexical representations or contextualized representations in isolation. Forcing a decoder to regress these isolated targets creates a severe representation gap due to inevitable information loss. To facilitate a systematic analysis, we formalize these targets into a dual-component configuration encompassing static linguistic attributes and dynamic contextual information. From the static perspective, mainstream embedding models map individual words to fixed vector spaces based on text co-occurrence statistics \cite{mikolov2013efficient,mikolov2013distributed,pennington2014glove}. While these static features capture stable conceptual coordinates, they represent words in complete isolation and omit essential contextual information \cite{xiao2024comparing}, leading to semantic ambiguity over continuous text. Conversely, the dynamic contextual component extracts representations from pre-trained language models \cite{devlin2018bert,radford2019language} to track long-range contextual sequences \cite{schrimpf2021neural,caucheteux2022brains,bowang2024icasspw}. Yet, relying solely on contextualized representations also introduces information loss, as these history-dependent embeddings often blur sharp local word boundaries and dilute the stable baseline meaning of individual words, allowing high-noise neural recordings to easily perturb discrete word retrieval.

This single-feature formulation directly deviates from the human brain's language processing mechanisms. Neurobiological evidence suggests that the human brain avoids this information loss by processing continuous language via two synergistic pathways: a lexical memory retrieval pathway that retrieves stable word attributes from distributed temporal networks \cite{lau2008cortical}, and a contextual unification pathway that dynamically integrates these units within fronto-temporal networks \cite{hagoort2005broca,hagoort2013fpsyg,hagoort2019science}. During continuous speech comprehension, the human brain simultaneously tracks stable lexical properties and contextual information to resolve semantic ambiguity \cite{sereno2003context,lau2008cortical,jain2018incorporating,pylkkanen2019neural,toneva2022combining,goldstein2022shared}, interacting closely within the classic N400 time window \cite{kutas2011thirty}.

Consequently, relying on a single isolated feature track is insufficient to capture the collaborative functions of human language networks. To better match the biological language processing mechanisms, a robust semantic reconstruction framework requires a multi-feature fusion approach. By forcing static lexical representations and contextualized representations to modulate each other to synthesize a unified target semantic feature space, the framework can prevent information loss and facilitate accurate cross-modal mapping.

\subsection{Proposed Framework and Core Contributions}
Crucially, the engineering objective of designing a unified semantic feature space is to improve the computational feasibility of neural decoding, rather than mechanically replicating biological brain structures. Since non-invasive neural signals (such as MEG and EEG) are highly non-stationary and extremely noisy, directly mapping them onto unconstrained language spaces can easily cause severe overfitting. This vulnerability is further exacerbated by the time-series autocorrelations inherent in non-invasive neural approaches \cite{xu2026bspc}. A multi-feature semantic space provides essential regularization; by constraining the decoding target within this well-defined representational space, we structurally narrow the hypothesis space. This constraint enables the decoding model to stably converge onto meaningful semantic targets despite the inherent noise of the neural data.

To implement this target semantic feature space, this study introduces a multi-feature fusion framework for semantic reconstruction. Departing from traditional single-feature approaches, we systematically formulate and contrast two alternative fusion methods to synthesize a unified semantic feature space: a linear fusion baseline via Naive Concatenation and a non-linear fusion driven by Multi-Head Cross-Attention. Within this cross-attention method, the static lexical representations and dynamic contextual information interactively attend to one another, simulating the complementary mechanisms of collaborative multi-feature processing native to human language networks.

Evaluated cross-modally on a continuous auditory narrative MEG dataset using self-supervised contrastive alignment, our framework yields three primary contributions:
\begin{itemize}
    \item \textbf{Dual-Feature Fusion Framework:} We propose a hybrid semantic feature target framework that integrates both static lexical properties and contextual information, improving the overall performance of both semantic reconstruction and continuous text generation.
    \item \textbf{Support for Neuroscientific Hypotheses:} By benchmarking linear and non-linear feature integration, our experiments reveal a robust performance hierarchy $(\text{Cross-Att} > \text{Concat} > \text{GPT} > \text{W2V})$. The quantitative supremacy of reciprocal cross-attention over naive concatenation provides support aligning with neuroscientific hypotheses regarding multi-feature cooperative processing during language comprehension.
    \item \textbf{Contrastive Alignment Method:} We implement a self-supervised contrastive learning method to optimize the cross-modal framework. This method effectively mitigates the difficulties inherent in semantic reconstruction using low SNR non-invasive neural signals.
\end{itemize}

\begin{figure}[t]
\centering
\includegraphics[width=1\textwidth]{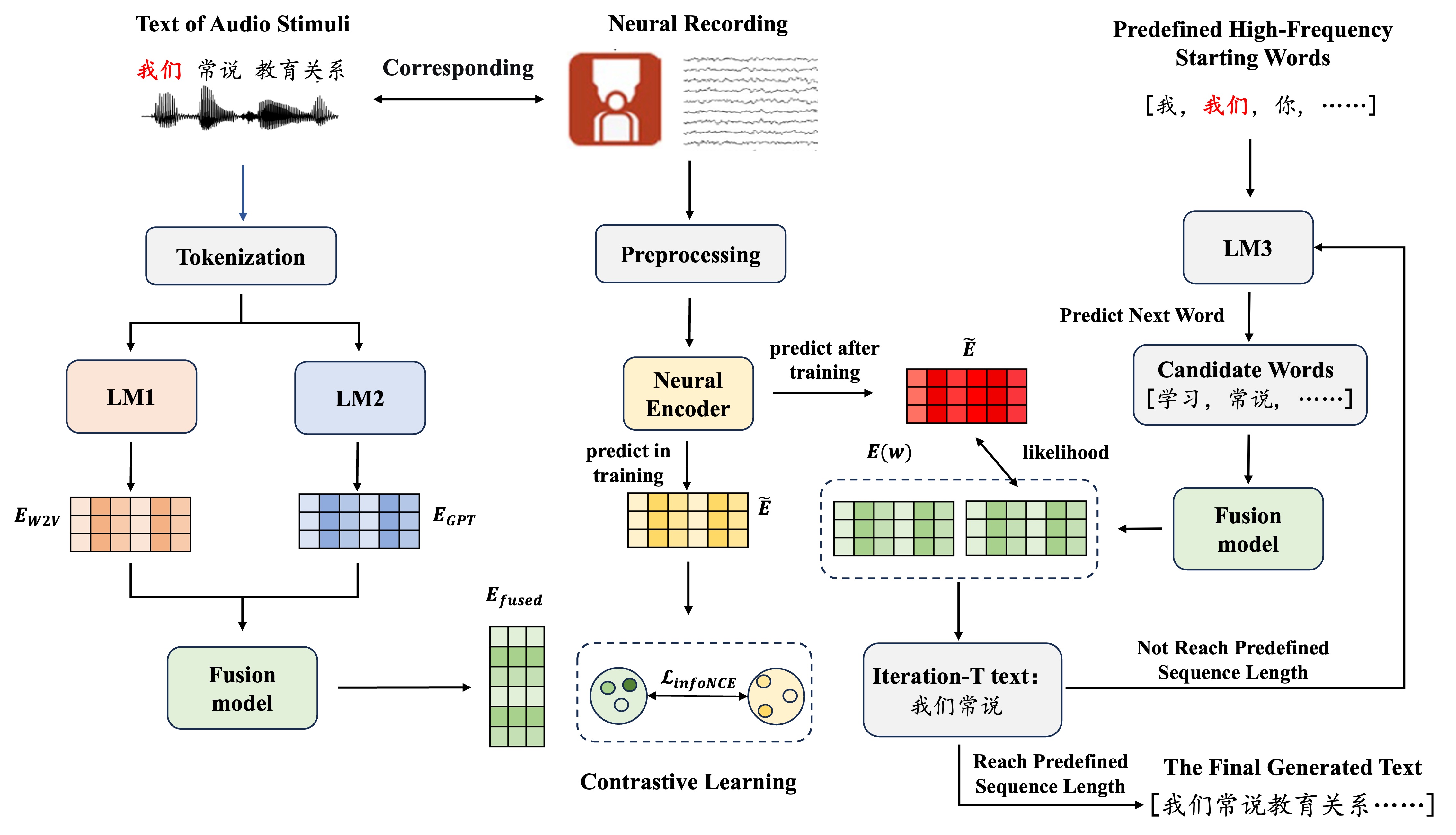}
\caption{\textbf{Global architecture of the proposed interactive Multi-feature fusion framework for non-invasive semantic reconstruction.} The method is structured as a two-stage framework: Stage 1: Semantic Feature Reconstruction and Contrastive Learning. Text of audio stimuli are processed via tokenization to feed LM1(Word2vec) that captures stable lexical attributes and LM2(GPT-Neo) that tracks dynamic sequence context. These representations are integrated by the Fusion model to get fused semantic features $E_{\text{fused}}$. Concurrently, neural recordings are mapped via a Neural Encoder into predicted fusion embedding. A batch-wide pairwise similarity matrix is constructed, governed by an InfoNCE loss function to maximize diagonal matches while minimizing off-diagonal distractors. Stage 2: Text Generation. Conditioned on the historical prefix and the fusion embeddings predicted from Stage 1, pre-trained LM3(mengzi-GPT-Neo) expands path hypotheses to generate candidate words. Each candidate word is sequentially transformed into its fusion embedding via Fusion model. The framework decides the next word by computing likehood between these candidate fusion embeddings and the predicted fusion embedding to output fluent, natural language text.}
\label{fig:framework}
\end{figure}

\section{Method}
In this section, we present the comprehensive methodology of our interactive Multi-feature fusion framework for non-invasive semantic reconstruction. The global architecture of the proposed framework is illustrated in Fig.~\ref{fig:framework}. The subsequent subsections will elaborate on the specific implementation and operational details of each component within the proposed framework.
\subsection{Methodological Overview and Design Motivations}
Before formalizing the specific matrix operations, we outline the structural hypotheses governing our framework. To model the dual facets of human language  processing (static lexical attributes and contextual information), we utilize two distinct semantic feature extractors to construct a hybrid target space. Crucially, we evaluate two opposite integration engineering methods: (1) \textit{Linear Concatenation Fusion}, which hypothesizes that static lexicon items and context sequence tracks are processed via isolated, passive parallel channels, and (2) \textit{Interactive Cross-Attention Fusion}, which introduces active, reciprocal non-linear gating blocks to let the two features dynamically modulate each other's latent geometries. These multi-tier targets are then mathematically aligned with high-dimensional neural time series using self-supervised contrastive learning to drive downstream text generation methods.

\subsection{Dataset}
\textbf{SEM4Lang.} The SEM4Lang dataset comprises 60.7 hours of magnetoencephalography (MEG) recordings gathered from 12 native Chinese speakers during a continuous speech listening task \cite{wang2022synchronized}. Throughout the experiment, participants listened to 60 distinct audio stories, each spanning 4 to 7 minutes and encompassing a diverse range of topics.  Neural activity was captured at a sampling rate of 1,000 Hz using a 306-channel whole-head MEG system (Elekta-Neuromag TRIUX, Helsinki, Finland).

\subsection{Semantic Feature Extraction and Preprocessing}
\textbf{Semantic Feature Extraction.} Sixty Chinese audios were utilized as continuous natural narrative speech stimuli. For the dynamic feature stream, we implemented a pre-trained Chinese GPT-Neo base model (\texttt{mengzi-GPT-Neo-base}\cite{langboat2021mengzi}, with a maximum sequence length $ML=128$ and a batch size $BS=64$). The hidden states of the 9th Transformer layer were extracted word-by-word utilizing a symmetrical contextual look-back window of 5 tokens ($\text{GPT\_WORDS}=5$). Given a tokenized sequence $w_1, w_2, \dots, w_n$, the dynamic contextual vector for the $i$-th token is formalized as:
\begin{equation}
e_{i^{\text{GPT}}} = \text{GPT-Neo-layer-9}([w_{i-5}, \dots, w_i])[-1, :] \in \mathbb{R}^{768}
\end{equation}

Subsequent to the dynamic method, for the static feature of the $i$-th token, 300-dimensional word-level representation $e_{i^{\text{W2V}}}$ is extracted using a Continuous Bag-of-Words (CBOW) Word2Vec framework trained over a Chinese text corpus using a context window of 10 \cite{mikolov2013distributed,mikolov2013efficient}. After that, based on the start and end times of each token, its dynamic and static features are tiled across the entire time interval it occupies, yielding the final representations $E_{\text{GPT}}\in \mathbb{R}^{T \times 768}$ and $E_{\text{W2V}}\in \mathbb{R}^{T \times 300}$, where $T$ represents the total number of sampling points of the entire story. Both $E_{\text{GPT}}$ and $E_{\text{W2V}}$ feature sequences were downsampled to 40 Hz using linear interpolation to achieve exact temporal alignment with the neural signals.

\subsubsection{MEG Signal Preprocessing.} MEG recordings collected across 12 healthy native Chinese speakers were preprocessed following standard artifact-rejection methods. For core regression mapping, only the 204 planar gradiometers were retained due to their superior signal-to-noise ratio (SNR) for localized cortical activity. Signals underwent spatial denoising (tSSS), independent component analysis (ICA)\cite{delorme2004eeglab,makeig1996independent}, lowpass filtering (4 Hz), and were downsampled to 40 Hz. The preprocessing was implemented with MNE-Python\cite{gramfort2014mne}.

\subsection{Semantic Feature Fusion}
\begin{figure}[H]
\centering
\includegraphics[width=1.0\textwidth]{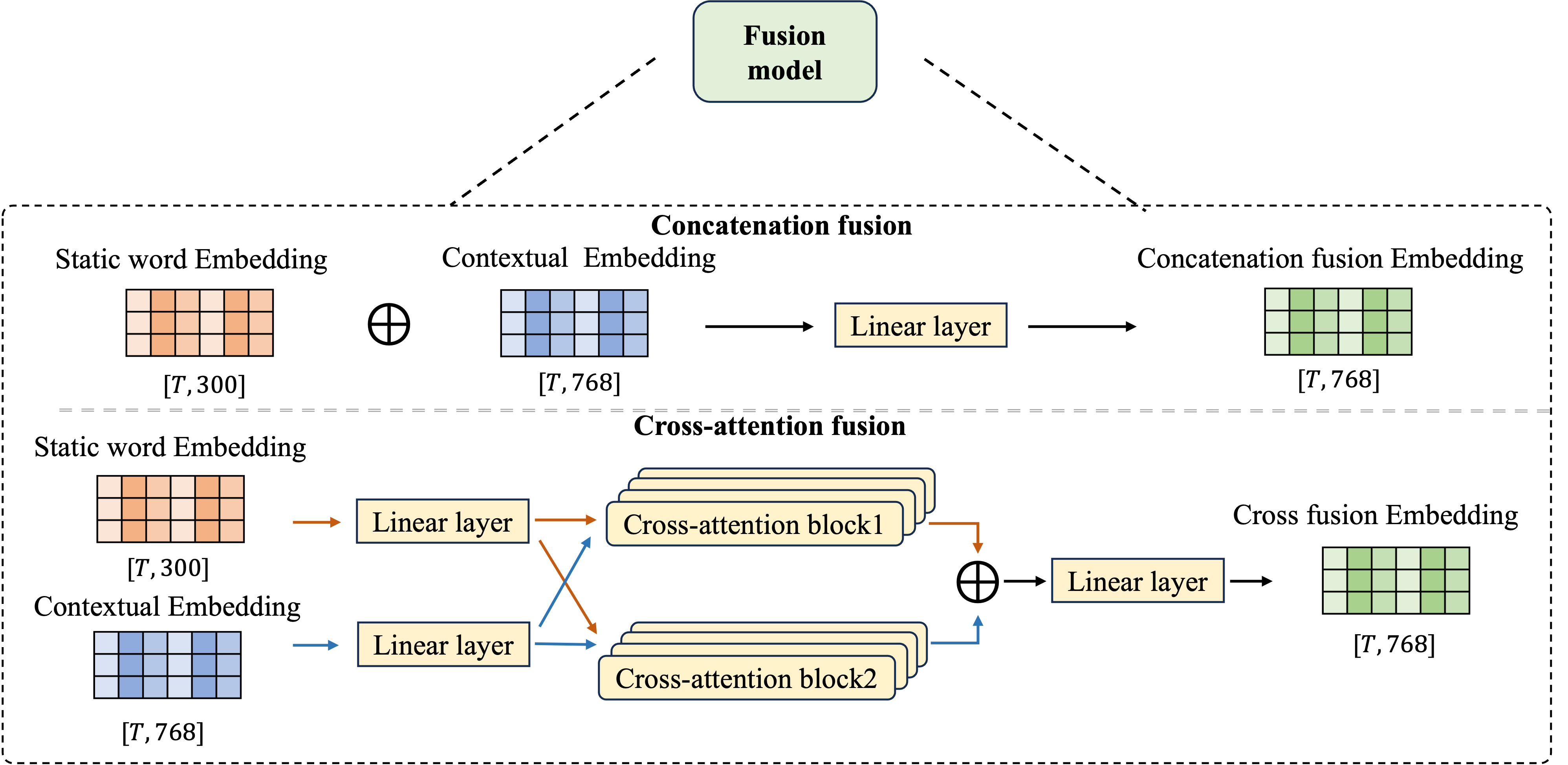}
\caption{The architectural topology of the proposed feature fusion methods. The upper block details the Concatenation Fusion baseline, where static vectors ([T,300]) and dynamic embeddings ([T,768]) are directly channel-concatenated and linearly projected.The lower block illustrates the Cross-Attention Fusion framework, where individual
linear layers first project both streams before routing them into reciprocal cross-attention
blocks.}
\label{fig:fusion_model}
\end{figure}

The overall detailed computational workflows for both the linear joint-feature baseline and the advanced non-linear interactive network are schematically illustrated in Fig.~\ref{fig:fusion_model}.

\subsubsection{Concatenation Fusion.} Our first proposed method represents a joint-feature baseline. It simply concatenates the static Word2Vec vector $E_{\text{W2V}} \in \mathbb{R}^{T \times 300}$ and the dynamic GPT vector $E_{\text{GPT}} \in \mathbb{R}^{T \times 768}$ directly along the channel dimension. While this baseline provides the brain decoder with access to both sources of information, it naively assumes that the human brain processes vocabulary and context through isolated parallel tracks without any non-linear cross-modal interaction. Furthermore, after feature fusion, we apply a linear projection layer to unify the concatenated dimensions to 768, matching the input requirements of the downstream decoder.

\subsubsection{Cross-Attention Fusion}
To simulate the highly non-linear, interactive feature synthesis native to human language unification, we introduce an advanced segment-level multi-modal fusion network (\texttt{FusionModel}) driven by a Multi-Head Cross-Attention block\cite{wei2020cross,vaswani2017attention,radford2021learning,lu2019vilbert}. The detailed topological flow is formalized as follows:

\noindent\textit{Projection and Reshaping.} The input vectors $E_{\text{W2V}}$ and $E_{\text{GPT}}$ are passed through individual Layer Normalization layers and linear projection layers to be mapped onto the target hidden dimension $D_{\text{out}}$. They are then reshaped into a token sequence format of $n_{\text{tokens}} \times d_{\text{token}}$ (where $d_{\text{token}} = D_{\text{out}} / n_{\text{tokens}}$ and $n_{\text{tokens}}$ is a tunable hyperparameter) to get the feature matrices $H_\text{GPT}$ and $H_\text{W2V}$:
\begin{equation}
H_\text{GPT} = \text{Reshape}(\text{LayerNorm}(\text{Linear}(E_{\text{W2V}}))) \in \mathbb{R}^{T \times n_{\text{tokens}} \times d_{\text{token}}}
\end{equation}
\begin{equation}
H_\text{W2V} = \text{Reshape}(\text{LayerNorm}(\text{Linear}(E_{\text{GPT}}))) \in \mathbb{R}^{T \times n_{\text{tokens}} \times d_{\text{token}}}
\end{equation}

\noindent\textit{Cross-Attention Interaction.} The two feature matrices serve asymmetric roles as Queries, Keys, and Values to perform dynamic cross-gating and reciprocal filtering:
\begin{equation}
\text{Attn}_\text{GPT} = \text{MultiheadAttn}(H_\text{GPT}, H_\text{W2V}, H_\text{W2V})
\end{equation}
\begin{equation}
\text{Attn}_\text{W2V} = \text{MultiheadAttn}(H_\text{W2V}, H_\text{GPT}, H_\text{GPT})
\end{equation}

\begin{figure}[H]
\centering
\includegraphics[width=0.60\textwidth]{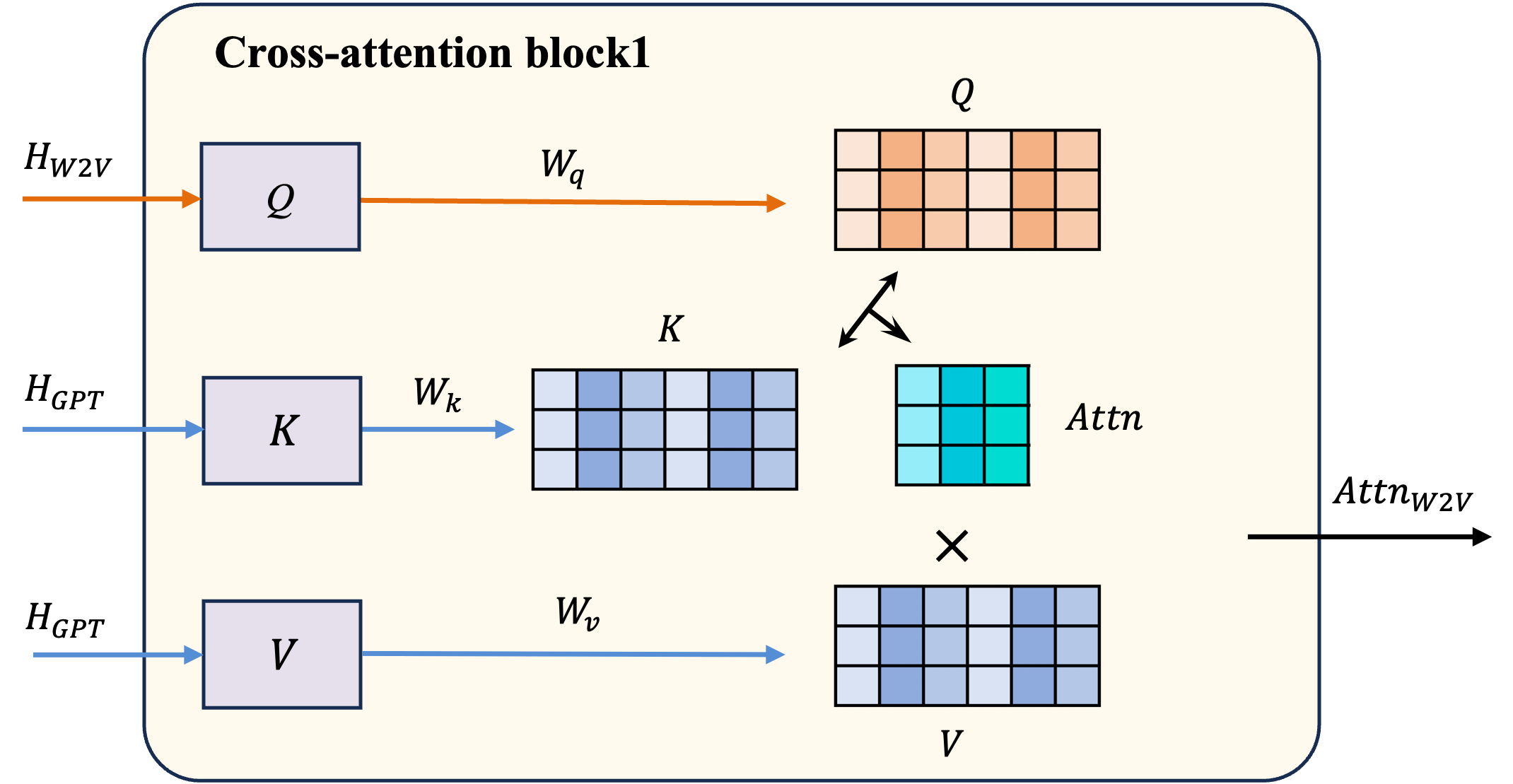}
\caption{The detailed computational method inside the Multi-Head Cross-Attention Block 1. Static lexical representations provide the structural query matrix $Q$ via projection weight $W_q$, while dynamic contextual vectors project into the corresponding key matrix $K$ and value matrix $V$ through $W_k$ and $W_v$. The scaled dot-product interaction between $Q$ and $K$ yields the dense attention matrix $Attn$. Finally, the value matrix $V$ is multiplied by the attention matrix $Attn$ to yield $Attn_{W2V}$.}
\label{fig:cross_attention_block1}
\end{figure}

The asymmetric cross-gating mechanism and step-by-step tensor routing inside the primary interactive cell are explicitly mapped out in Fig.~\ref{fig:cross_attention_block1}.

\noindent\textit{Fusion Output.} The bidirectional attention representations capturing complementary multi-level semantics are concatenated and passed through a Multi-Layer Perceptron (MLP) block, followed by temporal restoration:
\begin{equation}
E_{\text{fused}} = \text{MLP}([\text{Attn}_\text{GPT}; \text{Attn}_\text{W2V}]) \in \mathbb{R}^{T \times D_{\text{out}}}
\end{equation}
\subsection{MEG to Text}

The objective of this task is to directly decode and synthesize regular text narratives from non-invasive neural signals acquired during continuous story listening. As schematically outlined in Fig.~\ref{fig:framework}, this neural decoding method is formalized into a decoupled, two-stage cascade framework.

\subsubsection{Semantic Feature Reconstruction.} Our primary objective is to evaluate the model's capability to reconstruct comprehensive, multi-tier semantic language spaces directly from non-invasive neural signals. Rather than performing sound-level or wave-level acoustic matching, this task focuses entirely on semantic feature mapping and token-level lexical retrieval via the encoder network (\texttt{BrainNetwork}, implementing \texttt{ConvConcatNet}\cite{xu2024icasspw}, \texttt{BrainMagic}\cite{defossez2023decoding}, \texttt{AwaveNet}\cite{li2024cross,vandyck2023decoding}, \texttt{VLAAI}\cite{accou2023decoding}). The mathematical alignment method and the batch-wide contrastive optimization matrix used to bind the neural response embeddings with the interactive language targets are systematically detailed in Fig.~\ref{fig:framework}.

\noindent\textit{Dataset Partitioning.} To evaluate zero-shot generalization\cite{palatucci2009zero}, the 60 trials from the SEM4Lang dataset are randomly split into a training set of 40 trials, a validation set of 10 trials, and a test set of 10 trials. Continuous data are sliced into uniform, non-overlapping intervals with an epoch duration of $seglen = 3\text{s}$ across all subject sessions.

\noindent\textit{Self-Supervised Contrastive Learning and Loss Function.} The alignment method binds the neural response embeddings with the interactive multi-tier language targets. To capture exact temporal synchronization across the time series, the network minimizes a batch-wide InfoNCE contrastive loss\cite{chen2020simple,chen2021exploring} based on a pairwise temporal Pearson correlation similarity matrix. Within a mini-batch of size $B$, let $\tilde{E}_i$ represent the brain encoder's prediction for the $i$-th sample of the batch and $E_{\text{fused}, j}$ denote the true fused target for the $j$-th sample of the batch. The time-step wise Pearson similarity coefficient $(i, j)$ over the continuous temporal duration $T$ ($T = seglen \times 40$) is formalized as:
\begin{equation}
sim(i, j) = \sum_{t=1}^{T} \frac{(\tilde{E}_{i,t} - \overline{\tilde{E}}_{i,t})(E_{j,t}^{\text{fused}} - \overline{E}_{j,t}^{\text{fused}})}{(D_{\text{out}} - 1) \cdot \sigma(\tilde{E}_{i,t}) \cdot \sigma(E_{j,t}^{\text{fused}})}
\end{equation}
where $\overline{E}$ and $\sigma(\cdot)$ denote the mean and standard deviation of the feature vector along the channel dimension, respectively. Using this pairwise similarity metric, the batch-wide contrastive InfoNCE loss function $\mathcal{L}$ is constructed as:
\begin{equation}
\mathcal{L} = -\frac{1}{B} \sum_{i=1}^{B} \log \frac{\exp(sim(i, i))}{\sum_{j=1}^{B} \exp(sim(i, j))}
\end{equation}
By maximizing the diagonal scores $sim(i, i)$ and compressing the off-diagonal entries $sim(i, j)$, the loss forces the networks to map true brain-text associations together while pushing away mismatched negative distractors within the same batch.

\subsubsection{Text Generation.} This phase reconstructs continuous natural language word-by-word from neural signals.

Conditioned on the reconstructed neural embedding tensors $\tilde{E}$ from Stage 1, a heuristic beam search algorithm synthesizes text strings across discrete word timestamps $t_1, t_2, \dots, t_M$ over a deduplicated decoding vocabulary $\mathcal{V}$ ($3,535$ tokens), which merges high-frequency Chinese characters ($\ge 2$ occurrences in the first 50 stories) and the \texttt{mengzi-GPT-Neo} tokenizer. 

For each candidate token $w$, its virtual fused representation is obtained via:
\begin{equation}
E(w) = \text{fusion}(e_{\text{W2V}}(w), h_{\text{GPT}}(w))
\end{equation}
where $h_{\text{GPT}}(w)$ is the hidden state vector from the 9th Transformer layer computed by appending $w$ to the current path prefix inside the \texttt{Generation LM} under a contextual window of $\text{GPT\_WORDS} = 5$. The static vector $e_{\text{W2V}}(w)$ is indexed via a word look-up matrix $\mathcal{J}: \Sigma \rightarrow \mathbb{R}^{300}$.

To balance efficiency and fluency, the framework deploys a brain-guided heuristic beam search with a fixed beam width $B=100$ and step expansion parameter $E=5$. The recursive iteration proceeds as follows:
\begin{enumerate}
    \item \textit{Initialization:} The search tree is seeded with $E$ high-probability root nodes from the narrative starter pool in the tracking beam.
    \item \textit{Temporal Prior Extension:} For each word index $k = 2, 3, \dots, M$, the preceding tokens of each surviving path $h \in \mathcal{B}_{k-1}$ are fed into the \texttt{Generation LM} (bounded by a lower floor of 5 tokens and an upper threshold of $\text{LM\_TIME} = 8$ seconds) to output the conditional probability $P_{\text{LM}}(w|h_{\text{ctx}})$. Top-$K$ sampling ($top\text{-}k=10$) isolates a reduced candidate list $\mathcal{C}_h$ while filtering repetitive stop-words.
    \item \textit{Physiological Correlation Scoring:} Each candidate $w \in \mathcal{C}_h$ is converted into $e(w)$ and cross-matched against the reconstructed neural response segment $\tilde{E}[t_k, t_{k+1}]$ from Stage 1. The brain-responsive matching score is derived via cross-time average Pearson correlation coefficients:
    \begin{equation}
    s_{\text{brain}}(w) = \frac{1}{t_{k+1}-t_k} \sum_{t=t_k}^{t_{k+1}} \text{Pearson}(E(w), \tilde{E}[t])
    \end{equation}
    \item \textit{Adaptive Pruning:} Paths are allocated an expansion budget $n_h = |E \cdot \frac{\text{rank}(b)}{|\mathcal{B}_{k-1}|}|$ based on their accumulated LM log-probabilities. All expanded child hypotheses are pooled and sorted globally in descending order of $s_{\text{brain}}(w)$, preserving only the top $B$ paths:
    \begin{equation}
    \mathcal{B}_k = \text{Top-}B_{w \in \cup_h \mathcal{C}_h} (s_{\text{brain}}(w))
    \end{equation}
\end{enumerate}

\subsection{Evaluation}

\subsubsection{Semantic Feature Reconstruction.} The reconstruction fidelity is quantitatively evaluated using a Top-$K$ Rank Accuracy metric\cite{kay2008identifying}. All sub-segments from the testing set are cross-matched against the complete reference pool based on their temporal Pearson similarity scores. The candidates are sorted in descending order according to their similarity coefficients. A retrieval instance is classified as successful if the matched ground-truth semantic vector falls within the top $K$ positions of the sorted candidate array, where $K$ is 1 or 10.

\subsubsection{Text Generation.} The final text generation outputs are cross-matched against the ground-truth text transcripts at a character level across local temporal sliding windows ($\text{WINDOW} = 20$ seconds) to evaluate structural and semantic decoding stability over long narrative durations. The translation and language similarity benchmarks comprise: \textbf{BLEU-1} (measures token precision via unigram exact matches)\cite{papineni2002bleu}, \textbf{METEOR} (evaluates structural generation alignment by calculating comprehensive precision and recall scores)\cite{banerjee2005meteor}, and \textbf{BERTScore} (computes non-linear contextual semantic similarity leveraging layer-8 hidden state recall values from a pre-trained \texttt{bert-base-chinese} model)\cite{zhang2019bertscore}. 

To ensure that the text generation performance is heavily driven by neural grounding rather than language model hallucinations, we establish a \textbf{Null Baseline}\cite{combrisson2015exceeding}. At each forward token iteration, the real cortical correlation score $s_{\text{brain}}(w)$ is blocked and replaced with a uniform random variable $\sim \mathcal{U}(0, 1)$, keeping all other search and pruning parameters identical \cite{xu2026bspc}. Comparing the authentic fused performance against this opportunity baseline verifies the statistical significance of our non-invasive neuro-decoding framework.

\section{Results}
\subsection{Semantic Feature Reconstruction}
\begin{figure}[H]
\centering
\includegraphics[width=\textwidth]{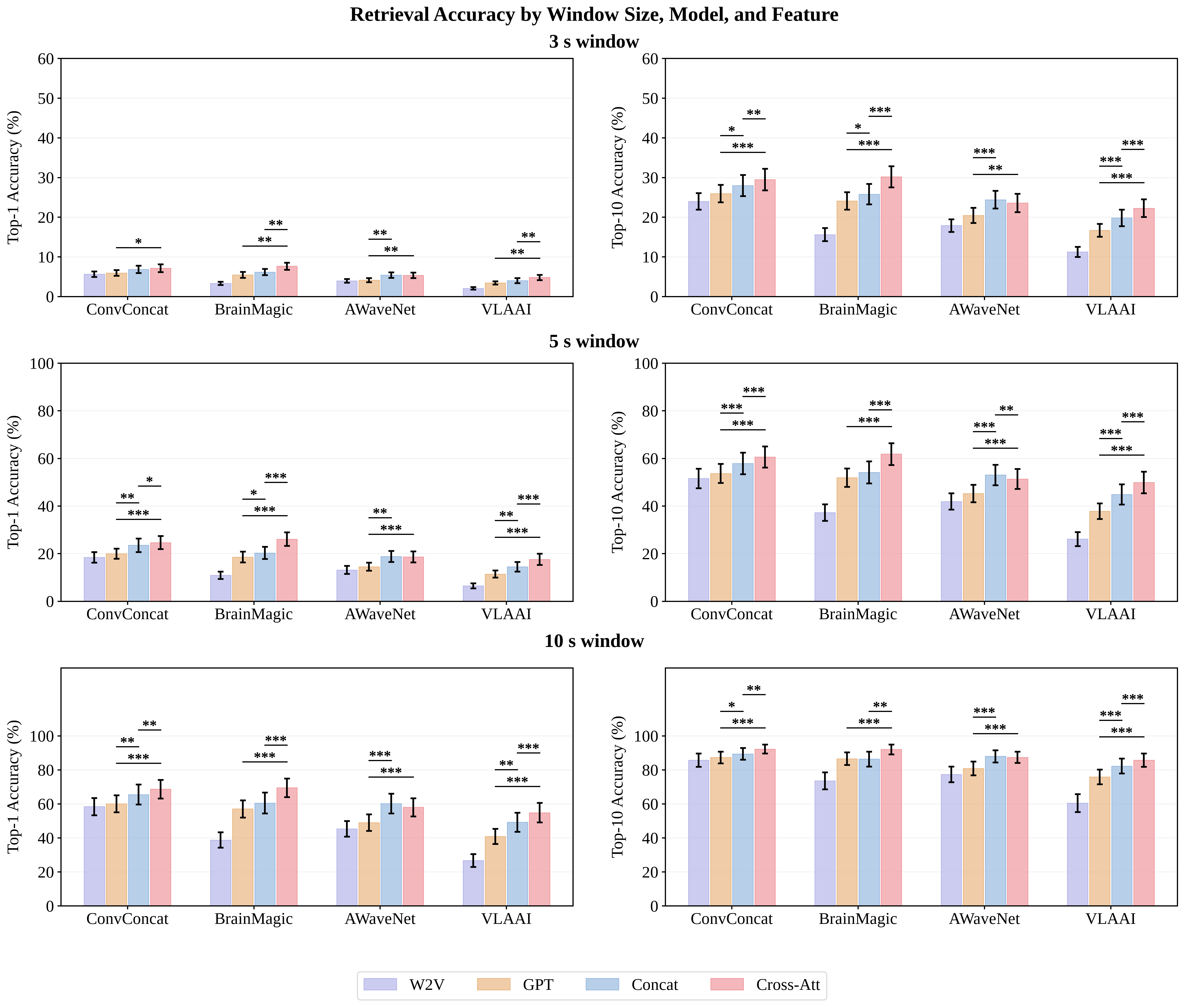}
\caption{Quantitative evaluation of semantic reconstruction performance measured by Top-1 (left column) and Top-10 (right column) retrieval accuracies across varying temporal window sizes (3\,s, 5\,s, and 10\,s). Four specialized neural decoders (\texttt{ConvConcat}, \texttt{BrainMagic}, \texttt{AWaveNet}, and \texttt{VLAAI}) are benchmarked across four linguistic and contextual configurations: static word embeddings (\text{W2V}), standalone contextual representations (\text{GPT}), linear joint features (\text{Concat}), and the proposed non-linear fusion method (\text{Cross-Att}). Bar heights represent the mean decoding accuracy, and error bars denote the standard error of the mean (SEM) computed across subject sessions ($n=12$). Horizontal lines indicate statistical significance derived from paired one-tailed $t$-tests ($^{*}\,p < 0.05$, $^{**}\,p < 0.01$, $^{***}\,p < 0.001$).}
\label{fig:retrieval_accuracies_full}
\end{figure}

We baseline four neural encoders (\texttt{ConvConcatNet} \cite{xu2024icasspw}, \texttt{BrainMagicNet} \cite{bowang2024ieee}, \texttt{AWaveNet} \cite{li2024cross}, and \texttt{VLAAI} \cite{accou2023decoding}) across four distinct linguistic and contextual configurations under a zero-shot semantic retrieval task(Fig.~\ref{fig:retrieval_accuracies_full}).

\noindent\textbf{Performance Hierarchy and Complementary Integration.} The empirical retrieval metrics reveal a robust and consistent performance hierarchy across all neural decoders and temporal segments, adhering to the inequality chain: $\text{Cross-Att} > \text{Concat} > \text{GPT} > \text{W2V}$ (Fig.~\ref{fig:retrieval_accuracies_full}). Specifically, the non-linear fusion method (\text{Cross-Att}) significantly outperforms the linear joint-feature baseline (\text{Concat}) across multiple time windows ($^{*}\,p < 0.05$, $^{**}\,p < 0.01$, $^{***}\,p < 0.001$), while both unified approaches surpass the standalone single-component spaces. This substantial gain demonstrates that complements between static linguistic attributes and dynamic contextual information can effectively elevate decoding capacity. 

Compared to the passive linear concatenation baseline, the non-linear cross-attention model achieves significantly tighter lexical attribute clustering and higher cross-modal alignment accuracy. This performance gap empirically demonstrates that simulating the interactive integration and collaborative modulation between static lexical properties and contextual profiles allows the decoding framework to yield superior representational alignment with endogenous neural dynamics.

\noindent\textbf{Pronounced Temporal Slicing Effect.} The decoding metrics reveal a strong positive correlation between temporal context duration (\textit{seglen}) and retrieval robustness. While retrieval in short windows ($seglen = 3\,\text{s}$) remains highly challenging due to sparse contextual information and the inherently low SNR of non-invasive recordings, extending the continuous epoch length to $10\,\text{s}$ triggers a pronounced surge in Top-10 accuracy (Fig.~\ref{fig:retrieval_accuracies_full}). This indicates that aggregating longer neural time series effectively smooths out high-frequency biological noise, elevates the signal-to-noise ratio, and provides stable global semantic boundaries to alleviate cross-modal alignment difficulties \cite{xu2026bspc}.

\noindent\textbf{Neural Architecture Comparison.} Among the specialized neural encoders, \texttt{ConvConcatNet} and \texttt{BrainMagicNet} emerge as the top-performing encoders. Specifically, \texttt{BrainMagicNet} paired with the \text{Cross-Att} configuration achieves prominent retrieval ceilings in the $5\,\text{s}$ and $10\,\text{s}$ tasks, while \texttt{ConvConcatNet} exhibits remarkable stability across all feature configurations. This validates that multi-scale temporal convolutions excel at capturing stable, localized neural states from non-stationary neuroimaging dynamics \cite{xu2024icasspw}.

\subsection{Text Generation}
To evaluate end-to-end decoding proficiency, the top two neural encoders (\texttt{BrainMagicNet} and \texttt{ConvConcatNet}) were interfaced with the downstream text generation stage, quantified via character-level metrics: BLEU-1 \cite{papineni2002bleu}, METEOR \cite{banerjee2005meteor}, and BERTScore \cite{zhang2019bertscore} (Table~\ref{tab:text_generation_results_updated}).

\begin{table}[H]
\centering
\caption{Quantitative text generation results across different neural encoders and semantic feature configurations. Bold values indicate the top-performing setup (performance ceiling) within each metrics category.}
\label{tab:text_generation_results_updated}
\resizebox{\textwidth}{!}{%
\begin{tabular}{lcccccc}
\toprule
Feature / & \multicolumn{3}{c}{\textbf{BrainMagicNet}} & \multicolumn{3}{c}{\textbf{ConvConcatNet}} \\
\cmidrule(lr){2-4} \cmidrule(lr){5-7}
Model & BLEU-1 $\uparrow$ & METEOR $\uparrow$ & BERTScore $\uparrow$ & BLEU-1 $\uparrow$ & METEOR $\uparrow$ & BERTScore $\uparrow$ \\
\midrule
Chance level (Null) & 10.77 & 5.89 & 51.50 & 10.77 & 5.89 & 51.50 \\
$\text{W2V}$                & $11.44 \pm 1.03$ & $6.40 \pm 0.62$ & $51.46 \pm 0.47$ & $11.76 \pm 1.40$ & $6.85 \pm 1.01$ & $52.33 \pm 0.75$ \\
$\text{GPT}$ & $13.85 \pm 1.16$ & $7.79 \pm 0.71$ & $52.27 \pm 0.59$ & $13.47 \pm 1.40$ & $7.76 \pm 0.98$ & $52.68 \pm 0.56$ \\
$\text{Concat}$      & $14.06 \pm 1.38$ & $8.28 \pm 0.93$ & $53.23 \pm 0.42$ & $14.38 \pm 1.49$ & $8.40 \pm 1.49$ & $53.03 \pm 0.77$ \\
$\mathbf{\text{Cross-Att}}$ & $\mathbf{15.12 \pm 1.10}$ & $\mathbf{8.70 \pm 0.73}$ & $\mathbf{53.58 \pm 0.38}$ & $\mathbf{15.58 \pm 1.16}$ & $\mathbf{9.23 \pm 0.89}$ & $\mathbf{53.62 \pm 0.45}$ \\
\bottomrule
\end{tabular}%
}
\end{table}

\noindent\textbf{Configuration Hierarchy and Architectural Synthesis.} The superiority of our advanced brain-to-text method is directly grounded in these empirical generation evaluations. As detailed in Table~\ref{tab:text_generation_results_updated}, the \text{Cross-Att} framework consistently yields the highest generation quality, achieving a BLEU-1 score of $15.58 \pm 1.16$ and a METEOR score of $9.23 \pm 0.89$ when paired with \texttt{ConvConcatNet}, outperforming the Null Baseline and standalone single-component configurations by a wide, statistically significant margin. 

\texttt{ConvConcatNet} demonstrates a pronounced advantage over \texttt{BrainMagicNet} across all configurations. This performance gap suggests that multi-scale temporal convolutional frameworks excel at capturing stable, granular local neural coding patterns required for precise sequence generation. Furthermore, both decoders significantly surpass the Chance Level (Null) baseline, demonstrating that the downstream text generation performance is genuinely driven by actual cortical alignment and neural grounding, rather than by deep language model hallucinations.

\section{Conclusion}
\subsection{Summary and Implications}
This study introduced a multi-feature fusion framework to address the representational mismatch between computational semantic features and multi-channel non-invasive neural recordings. By transitioning from standalone, single-feature decoding targets to synthesized representation spaces, we formalized and benchmarked linear Naive Concatenation against a non-linear network governed by Multi-Head Cross-Attention.

Our semantic reconstruction and text generation experiments revealed a consistent performance hierarchy $(\text{Cross-Att} > \text{Concat} > \text{GPT} > \text{W2V})$ across multiple architectures, backed by statistical validation. The empirical success of the cross-attention method demonstrates that non-invasive language decoding benefits from simulating the interactive integration and collaborative modulation between static linguistic attributes and dynamic contextual information, establishing a target space that improves cross-modal alignment accuracy. This multi-feature synthesis establishes a target representation space that is computationally isomorphic to the neural coding patterns of the human language network. By constraining the decoding model within this cognitively grounded framework, we reduce the search space against the non-stationary noise floor inherent in non-invasive neuroimaging, thereby providing a robust method for continuous brain-to-text reconstruction.

\subsection{Future Work}
To transition this framework toward practical BCI deployment, future research will pursue two primary directions:
\begin{itemize}
    \item \textbf{Cross-Subject Generalization:} To mitigate prominent inter-subject variability, we aim to incorporate adversarial domain adaptation layers to align latent neural signals across subjects, thereby minimizing the calibration overhead for novel users.
    \item \textbf{Open-Source Benchmarking:} To foster reproducibility and collaborative development, we will release our multi-encoder encoders, preprocessing approaches, and text generation modules as a standardized, open-source Python library to provide a unified evaluation platform for the BCI community.
\end{itemize}

\bibliographystyle{splncs04}
\bibliography{copy_v1}

\end{document}